\pdfoutput=1

\documentclass[11pt]{article}

\usepackage[]{EMNLP2022}

\usepackage{times}
\usepackage{latexsym}

\usepackage[T1]{fontenc}

\usepackage[utf8]{inputenc}

\usepackage{microtype}

\usepackage{inconsolata}

\usepackage{soul}
\usepackage{url}
\usepackage{graphicx}
\usepackage{amsmath}
\usepackage{booktabs}
\usepackage{color}

\usepackage{amsfonts}
\usepackage{multirow}
\usepackage{colortbl}
\usepackage{pgfplots}
\usepackage{tikz}
\usepackage{courier}
\usepackage{pgfplotstable}
\usepackage{bbding}
\usepackage{framed}
\definecolor{shadecolor}{rgb}{0.92,0.92,0.92}
\usepackage{tcolorbox}
\usepackage{amssymb}
\usepackage{pifont}
\newcommand{\xmark}{\ding{56}}%
\usepackage{setspace}
\usepackage{makecell}
\usepackage{calligra}
\usepackage{algpseudocode} 
\usepackage{subfigure}
\usepackage{algorithm}

\title{Label-Driven Denoising Framework for Multi-Label Few-Shot \\Aspect Category Detection}

\author{
Fei Zhao$^{1*}$\quad Yuchen Shen$^2${\thanks{$^*$ Equal contributions.}}\quad Zhen Wu$^1${\thanks{\ \ \ Corresponding author.}}\quad Xinyu Dai$^1$ \\
$^1$National Key Laboratory for Novel Software Technology, Nanjing University\\
$^2$School of Information and Software Engineering, \\University of Electronic Science and Technology of China\\
{\tt zhaof@smail.nju.edu.cn, alexchicharitoshen@gmail.com}\\
{\tt \{wuz, daixinyu\}@nju.edu.cn}
}

\begin{document}
\maketitle

\begin{abstract}
Multi-Label Few-Shot Aspect Category Detection (FS-ACD) is a new sub-task of aspect-based sentiment analysis, which aims to detect aspect categories accurately with limited training instances. Recently, dominant works use the prototypical network to accomplish this task, and employ the attention mechanism to extract keywords of aspect category from the sentences to produce the prototype for each aspect. However, they still suffer from serious noise problems: (1) due to lack of sufficient supervised data, the previous methods easily catch noisy words irrelevant to the current aspect category, which largely affects the quality of the generated prototype; (2) the semantically-close aspect categories usually generate similar prototypes, which are mutually noisy and confuse the classifier seriously. In this paper, we resort to the label information of each aspect to tackle the above problems, along with proposing a novel Label-Driven Denoising Framework (LDF). Extensive experimental results show that our framework achieves better performance than other state-of-the-art methods.
\end{abstract}

\section{Introduction}

Aspect Category Detection (ACD) is an important subtask of fine-grained sentiment analysis~\cite{DBLP:conf/semeval/PontikiGPPAM14}, which aims to detect the aspect categories mentioned in a review sentence from a predefined set of aspect categories. For example, given the sentence ``\emph{The service is good although rooms are pretty expensive.}'', the ACD task is to detect two aspect categories from the sentence, respectively \emph{service} and \emph{price}. Obviously, the ACD belongs to a multi-label classification problem.

\begin{table}[t]
\centering
\scalebox{0.65}{
\begin{tabular}{p{4.2cm}|p{6.5cm}}
\toprule
\multicolumn{2}{c}{\textbf{Support set}} \\
\midrule
{Aspect Category} & {Sentences} \\ \hline
\multirow{4}{*}{(A) food\_food\_meat\_burger} & (1) {\it first time, burger was not fully cooked and my smash fries were cold.} \\ & (2) {\it food was over priced, but okay not great.} \\
\midrule
\multirow{3}{*}{(B) food\_mealtype\_lunch} & (1) {\it my brother and i stopped in for lunch.} \\ & (2) {\it lunch has a great option of picking one or two food with rice.} \\
\midrule
\multirow{2}{*}{(C) restaurant\_location} & (1) {\it i prefer the other location to be honest.} \\ & (2) {\it there's a new standard in town.} \\
\midrule
\multicolumn{2}{c}{\textbf{Query set}} \\
\midrule
Aspect Category & Sentences \\ \hline
(B) & (1) {\it went back today for lunch.} \\
(A) and (C) & (2) {\it food is whats to be expected at a neighborhood grill.} \\
\bottomrule
\end{tabular}}
\caption{An example of 3-way 2-shot meta-task. A sentence (instance) may belong to multiple aspects.}
  \label{tab:metatask}%
\end{table}


Recently, with the development of deep learning technique, a great number of neural models have been proposed for the ACD task \cite{DBLP:conf/aaai/ZhouWX15,DBLP:journals/tcyb/SchoutenWFD18,DBLP:conf/emnlp/HuZZCSCS19}. The performance of all these models heavily rely on sufficient labeled data. However, the annotation of aspect categories in ACD is extremely expensive. The limited labeled data restrict the effectiveness of neural models. To alleviate the issue, ~\citeauthor{DBLP:conf/acl/HuZGXGGCS20}~\shortcite{DBLP:conf/acl/HuZGXGGCS20} refer to few-shot learning (FSL) \cite{DBLP:conf/iclr/RaviL17,DBLP:conf/icml/FinnAL17,DBLP:conf/nips/SnellSZ17,DBLP:conf/aaai/GaoH0S19} and formalize ACD as a few-shot ACD (FS-ACD) problem, learning aspect categories with limited supervised data.

FS-ACD follows the meta-learning paradigm \cite{DBLP:conf/nips/VinyalsBLKW16} and builds a collection of $N$-way $K$-shot meta-tasks. Table \ref{tab:metatask} shows a 3-way 2-shot meta-task, which consists of a support set and a query set. The support set samples three classes (i.e., aspect categories), and each class selects two sentences (instances). A meta-task aims to infer the classes of sentences in the query set with the help of the small labeled support set. By sampling different meta-tasks in the training stage, FS-ACD can learn great generalization ability in few-shot scenario and works well in the testing stage. To perform the FS-ACD task, ~\citeauthor{DBLP:conf/acl/HuZGXGGCS20}~\shortcite{DBLP:conf/acl/HuZGXGGCS20} proposes an attention-based prototypical network \emph{Proto-AWATT}. It first exploits an attention mechanism ~\cite{DBLP:journals/corr/BahdanauCB14} to extract keywords from the sentences corresponding to aspect category in the support set, and then aggregate them as evidence to generate a prototype for each aspect. Next, the query set utilizes the prototypes to generate corresponding query representations. Finally, the prediction is made by measuring the distance between each prototype representation and corresponding query representation in the embedding space.

\begin{figure}[t]
\centering
\includegraphics[width=3.0in]{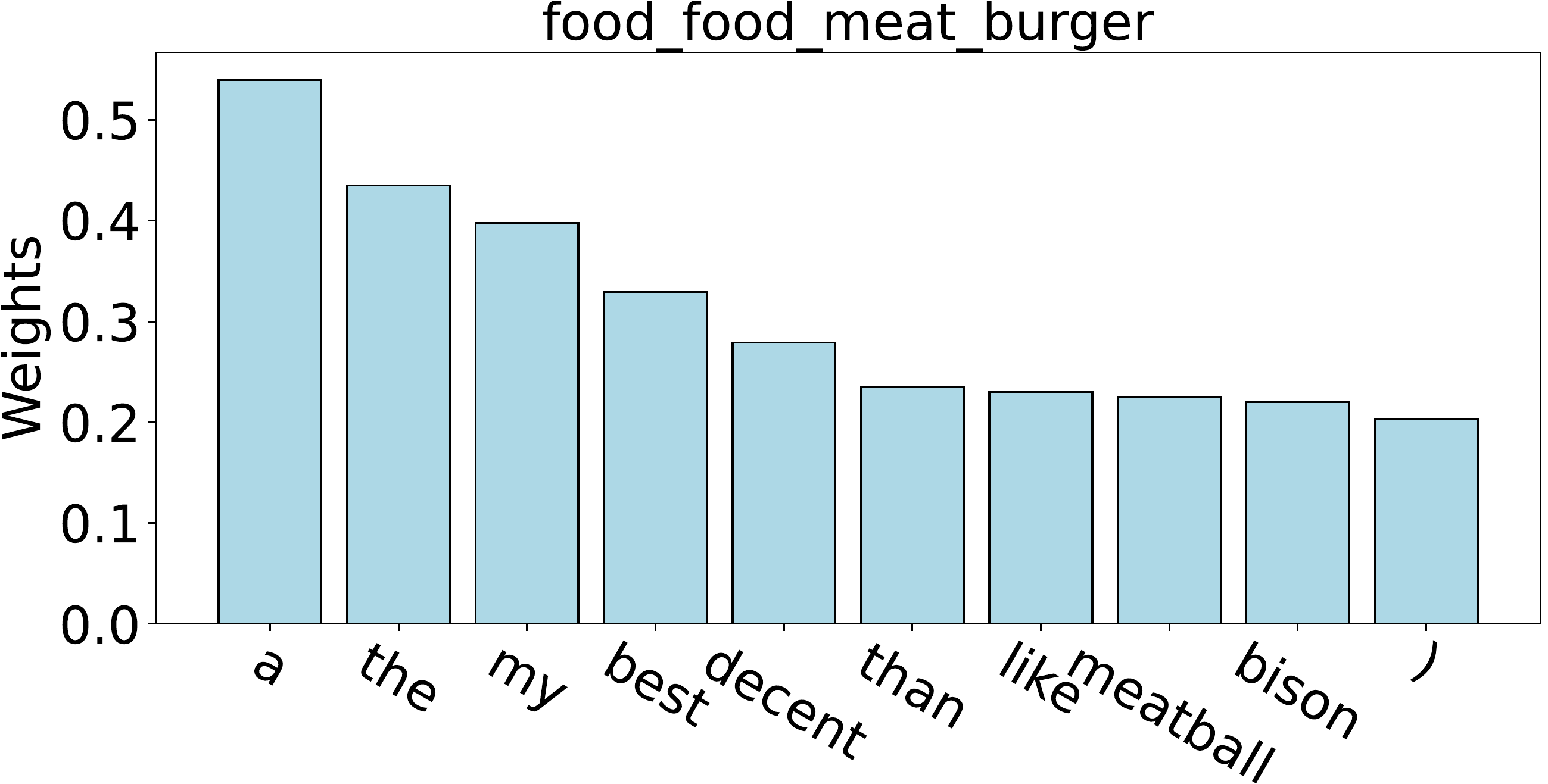}
\caption{Visualization of the top-10 words for the prototype of aspect category \emph{food\_food\_meat\_burger} according to the attention weights of \emph{Proto-AWATT}.}
\label{figure_a}
\end{figure}

Though achieving impressive progress, we find the noise is still a crucial problem for the FS-ACD task. The reason comes from two folds. \textbf{On the one hand}, the previous models easily catch noisy words irrelevant to the current aspect category due to the lack of sufficient supervised data, which largely affects the quality of the generated prototype. As shown in Figure~\ref{figure_a}, take the prototype of aspect category \emph{food\_food\_meat\_burger} as an example. We highlight its top-10 words based on attention weights of \emph{Proto-AWATT}. Because of lacking sufficient supervised data, we observe the model tends to focus on these common but noisy\footnote{we randomly sample 100 meta-tasks in the benchmark dataset and then visualize the top-10 words of each prototype in the support set based on the attention weight of \emph{Proto-AWATT}. According to the statistics, about 31.4\% of the prototypes assign the highest three attention weights to those common but noisy words.} words, such as ``a'', ``the'', ``my''. These noisy words fail to produce a representative prototype for each aspect, resulting in the discounted performance. \textbf{On the other hand}, the semantically-close aspect categories usually produce similar prototypes, these close prototypes are mutually noisy and confuse the classifier greatly. According to the statistics, nearly 25\%\footnote{We calculate the similarity between the embeddings of different aspect categories, and the similarity of semantically-close aspect category pairs is generally greater than 0.85.} of aspect category pairs in the benchmark dataset have similar semantics, such as \emph{food\_food\_meat\_burger} and \emph{food\_mealtype\_lunch} in Table~\ref{tab:metatask}. 
Apparently, the prototypes generated by these semantically-close aspect categories can interfere with each other and confuse the detection results of FS-ACD seriously.

To tackle the above issues, we propose a novel \textbf{L}abel-Driven \textbf{D}enoising \textbf{F}ramework (LDF) for the FS-ACD task. Specifically, for the first issue, the label text of aspect category contains rich semantics describing the concept and scope of aspect, such as the text ``\emph{restaurant location}'' for the aspect \emph{restaurant\_location}, which intuitively help the attention capture label-relevant words better. Therefore, we propose a label-guided attention strategy to filter noisy words and guide LDF to yield better aspect prototypes. Given the second issue, we propose an effective label-weighted contrastive loss, which incorporates inter-class relationships of support set into a contrastive objective function, thereby enlarging the distance among similar prototypes. 


Our contributions are summarized as follows:

\begin{itemize}
    \item To the best of our knowledge, we are the first to exploit the label information of each aspect to address noise problems in the FS-ACD task.
    \item We propose a novel Label-Driven Denoising Framework (LDF), which contains a label-guided attention strategy to filter noisy words and generate a representative prototype for each aspect, and a label-weighted contrastive loss to avoid generating similar prototypes for semantically-close aspect categories.
    \item The LDF framework has good compatibility and can be easily extended to existing models. In this work, we apply it to two latest FS-ACD models, \emph{Proto-HATT}~\cite{DBLP:conf/aaai/GaoH0S19} and \emph{Proto-AWATT}~\cite{DBLP:conf/acl/HuZGXGGCS20}. Experimental results on three benchmark datasets prove the superiority of our framework.
\end{itemize}


\section{Notations and Background}


In this section, we first present the task formalization of FS-ACD and then give brief introductions to the background.


\subsection{Task Formalization}

The FS-ACD task follows the meta-learning paradigm \cite{DBLP:conf/nips/VinyalsBLKW16}. Specifically, given labeled instances from a set of classes (i.e., aspect categories) $C_{train}$, the goal is to acquire knowledge from $C_{train}$ and use the knowledge to recognize novel classes, which have only a few labeled instances. These novel classes belong to a set of classes $C_{test}$ and disjoint from $C_{train}$.

To emulate the few-shot scenario, meta-learning algorithms learn from a group of $N$-way $K$-shot meta-tasks sampled from $C_{train}$. Within each meta-task, we randomly select $N$ classes ($N$-way) from $C_{train}$, each with $K$ instances ($K$-shot) to form a support set $\mathcal{S}=\{\mathbf{\emph{s}}_k^{n}\,|\,k=1,...,K\}_{n=1}^{N}$. Meanwhile, $M$ instances are sampled from the remaining data of the $N$ classes to construct a query set $\mathcal{Q}=\{(\mathbf{\emph{q}}_i,\,\mathbf{\emph{y}}_i)\,|\,\mathbf{\emph{y}}_i \in \mathbb{R}^{N}\}_{i=1}^{M}$, where $\mathbf{\emph{y}}_i$ is a binary label vector whose $n$-th bit is set to 1 if $\mathbf{\emph{q}}_i$ belongs to the $n$-th class (i.e., aspect category), 0 otherwise. A meta-task aims to infer the class(es) of query instance $\mathbf{\emph{q}}_i$ in $\mathcal{Q}$ according to a small labeled support set $\mathcal{S}$. By sampling different meta-tasks in the training stage, FS-ACD can learn great generalization ability. During the testing stage, we apply the same manner to test whether our model can adapt quickly to novel classes within $C_{test}$.

\subsection{Background}

In this work, we abstract a general attention architecture based on the \emph{Proto-AWATT}~\cite{DBLP:conf/acl/HuZGXGGCS20} and \emph{Proto-HATT}~\cite{DBLP:conf/aaai/GaoH0S19} models, which both achieve satisfying performance and thus are chosen as the foundations of our work.



Given a instance $\mathbf{\emph{s}}_k^{n}=\{w_1, w_2, ..., w_l\}$ consisting $l$ words, we first map it into an word sequence $\mathbf{\emph{e}}_k^{n}=\{\mathbf{\emph{e}}_1, \mathbf{\emph{e}}_2, ..., \mathbf{\emph{e}}_l\}$ by looking up an embedding table. And then, we apply a convolutional neural network (CNN) \cite{DBLP:conf/coling/ZengLLZZ14,DBLP:conf/aaai/GaoH0S19} to encode the word sequence into a contextual representation $H_k^n$. Next, an attention layer assigns a weight $\beta$ to each word in the instance. The final instance representation is given by:
\begin{gather}
\beta = \text{ATT}_W(\mathbf{\emph{H}}_k^n)\label{beta},\\
\mathbf{\emph{r}}_k^n = \beta \mathbf{\emph{H}}_k^n\label{representation},
\end{gather}
where $H_k^n$ is the $k$-th instance representation of the class $n$ in the support set $\mathcal{S}$, $\text{ATT}_W(\cdot)$ denotes an attention mechanism. After that, we aggregate all instance representations for the class $n$ to produce the prototype:
\begin{gather}
\mathbf{\emph{r}}^n=\text{Aggregation}(\mathbf{\emph{r}}_1^{n},...,\mathbf{\emph{r}}_K^{n}),\label{prototype}
\end{gather}
where $\text{Aggregation}(\cdot)$ denotes the attention mechanism or average pooling operation. After processing all classes in the support set $\mathcal{S}$, we obtain $N$ prototypes $\{\mathbf{\emph{r}}^1, \mathbf{\emph{r}}^2, ..., \mathbf{\emph{r}}^n, ..., \mathbf{\emph{r}}^N\}$.






Similarly, for a query instance $\mathbf{\emph{q}}_i$, we first encode $\mathbf{\emph{q}}_i$ to obtain its contextual representation, and then exploit an attention mechanism to produce $N$ prototype-specific query representations $\mathbf{\emph{r}}_i^n$ based on the $N$ prototypes. After that, we compute the Euclidean distance (ED) between each prototype and the corresponding prototype-specific query representation. Finally, we normalize the negative Euclidean distances to obtain the ranking of prototypes and use a threshold to select the positive predictions (i.e., aspect categories).
\begin{equation}
\hat{\mathbf{\emph{y}}_i} = \text{softmax}(-\text{ED}(\mathbf{\emph{r}}^n, \mathbf{\emph{r}}_i^n)), n\in [1, N]\label{ED}
\end{equation}

The training objective is the mean square error (MSE) loss as follows:
\begin{equation}
\mathcal{L}_{mse} = \sum_{i=1}^M(\hat{\mathbf{\emph{y}}_i} - \mathbf{\emph{y}}_i)^2
\end{equation}

\begin{figure*}[t]
\centering
\includegraphics[width=6.2in]{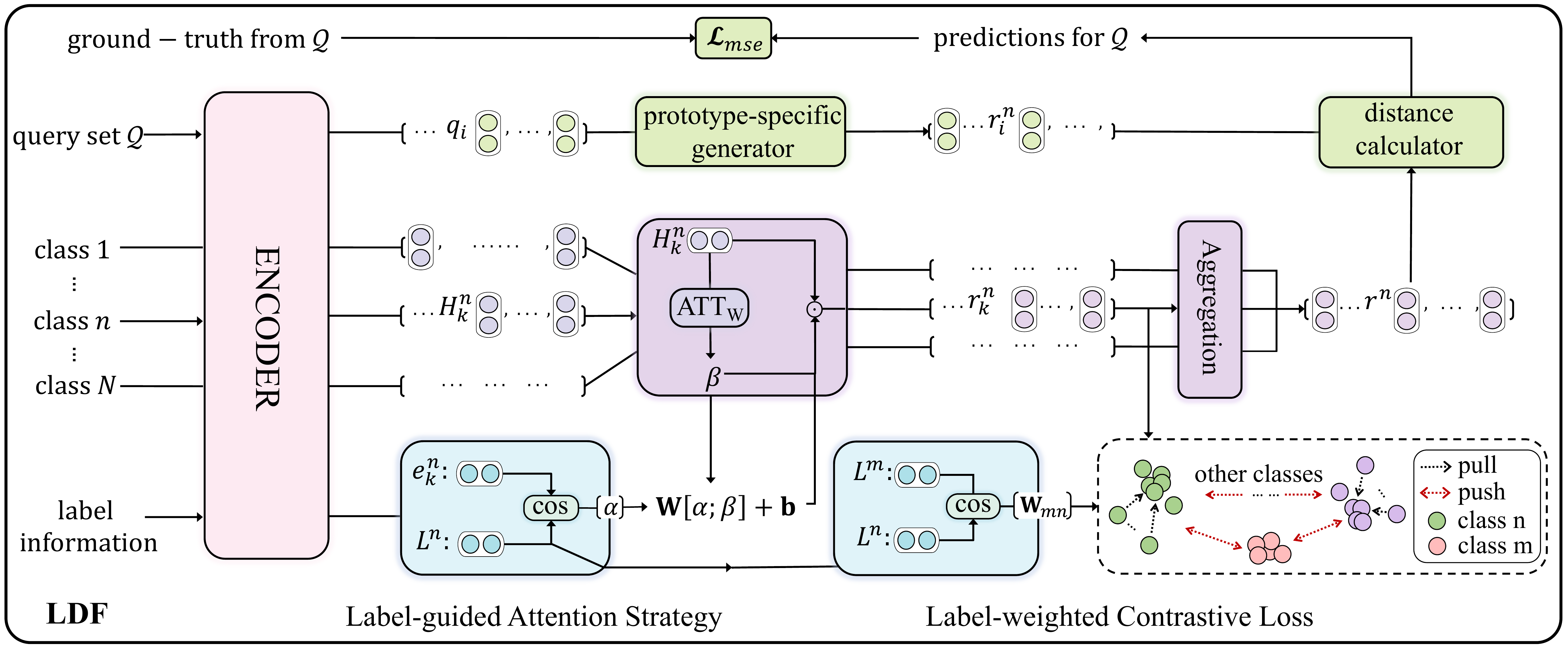}
\caption{The overview of our proposed LDF framework.}
\label{fig:overview}
\end{figure*}

\section{Label-Driven Denoising Framework}




Figure \ref{fig:overview} shows the overall architecture of LDF, which contains two components: Label-guided Attention Strategy and Label-weighted Contrastive Loss. With the aid of label information, the former can focus on the class-relevant words better, thus producing a more accurate prototype for each class, the latter utilizes the inter-class relationships of support set to avoid generating similar prototypes.




\subsection{Label-guided Attention Strategy}


Due to lack of sufficient supervised data, the attention weights $\beta$ in Equation \ref{beta} usually focus on some noisy words irrelevant to the current class (i.e., aspect category), resulting in the prototype in Equation \ref{prototype} becoming unrepresentative.





Intuitively, the label text of each class contains rich semantics, which can provide guidance for capturing class-relevant words. Thus, we leverage label information to tackle the above problem and propose a Label-guided Attention Strategy.







Specifically, we first locate the keywords of each class by calculating the semantic similarity between the label and each word in the instance:
\begin{equation}
\alpha = \text{cos}(\mathbf{\emph{L}}^n, \mathbf{\emph{e}}_k^n)\label{alpha},
\end{equation}
where $\mathbf{\emph{L}}^n$ is the label embedding of class $n$ in the support set and calculated by averaging the multiple word embeddings of each class (e.g., {\it food\_food\_meat\_burger}), $\mathbf{\emph{e}}_k^n$ is the word embedding of instance $\mathbf{\emph{s}}_k^{n}$, cos($\cdot$) is the cosine function.

Under the constraints of label information, the similarity weight $\alpha$ tends to focus on the limited words\footnote{We randomly sample 100 meta-tasks in the benchmark dataset and then visualize the words focused by each class in the support set based on the similarity weight $\alpha$. Statistically, around 79\% of the classes can only focus on less than 4 words each time, resulting in the prototype generated by them not being robust. Thus, we only use it as complementary information of the attention weight, instead of directly replacing the attention weight. The results in Table \ref{tab:alpha} also verify this point.} highly relevant to the label text and may neglect other informative words. Thus, we take it as the complementary information of the attention weights $\beta$ to generate a more comprehensive and accurate attention weight $\theta$. Formally,
\begin{equation}
\theta = W_g [\alpha; \beta] + b_g
\end{equation}
where $W_g$ and $b_g$ are weight matrices and bias, $[\cdot\,;\cdot]$ denotes the concatenation operation.

Then, to regain the probabilistic attention distribution, the attention weight $\theta$ is re-normalized:
\begin{equation}
\widetilde{\theta} = \textrm{softmax}(\theta)
\end{equation}

Finally, we replace $\beta$ in Equation \ref{beta} with the new attention vector $\widetilde{\theta}$ to obtain a representative prototype for each class in the support set.


\subsection{Label-weighted Contrastive Loss}

As mentioned before, the semantically-close aspect categories often generate similar prototypes in the support set, which are mutually noisy and confuse the classifier seriously.









Intuitively, a feasible and natural approach is to leverage supervised contrastive learning (CL) \cite{DBLP:conf/nips/KhoslaTWSTIMLK20}, which can push the prototype of different classes away as follows:
\begin{equation}
\resizebox{.97\hsize}{!}{$
\begin{split}
\mathcal{L}_{scl} &= \sum_{(n, k) \in (N, K)}\frac{-1}{|P(n, k)|}\\&\sum_{r_p^n \in P(n, k)}\text{log} \frac{\exp(r_k^n \cdot r_p^n / \tau)}{\sum_{r_k^m \in (N, K) \setminus (n, k)}\exp(r_k^n \cdot r_k^m / \tau)}
\end{split}$}
\end{equation}
where $P(n, k)$ is the positive set of $r_k^n$ in Equation \ref{representation}, which contains all the other samples (e.g., $r_p^n$) of the same class with $r_k^n$ in the support set. The rest of the ($N$-1)$\times K$ samples in the support set belong to the negative set, where $r_k^m$ is one negative sample from class $m$, $\tau$ is a temperature parameter.







However, the supervised CL does not well-resolve our problem since it treats different prototypes equally in the negative set, while our goal is to encourage the more similar prototypes to be farther apart. For example, ``{\it food\_food\_meat\_burger}'' is semantically closer to ``{\it food\_mealtype\_lunch}''  than ``{\it room\_bed}''. Thus, ``{\it food\_food\_meat\_burger}'' should be farther from ``{\it food\_mealtype\_lunch}'' than ``{\it room\_bed}'' in the negative set.

To achieve this goal, we again leverage the label information and propose to incorporate inter-class relationships into the supervised CL to adaptively distinguish similar prototypes in the negative set:
\begin{equation}
\resizebox{.97\hsize}{!}{$
\begin{split} &\mathcal{L}_{lcl} = \sum_{(n, k) \in (N, K)}\frac{-1}{|P(n, k)|}\\ &\sum_{r_p^n \in P(n, k)} \text{log} \frac{\exp(r_k^n \cdot r_p^n / \tau)}{\sum_{r_k^m \in (N, K) \setminus (n, k)}w_{mn}\cdot \exp(r_k^n \cdot r_k^m / \tau)}
\end{split}$}
\end{equation}
where $w_{mn}$ denotes the cos similarity between different classes in the negative set and is computed as follows:
\begin{equation}
w_{mn} = \text{cos}(L^m, L^n),
\end{equation}
where $L^m$ and $L^n$ are the label embedding of the class $m$ and $n$. The final loss is formulated as:
\begin{equation}
\mathcal{L} = \mathcal{L}_{mse} + \lambda\mathcal{L}_{lcl}
\end{equation}
where $\lambda$ is a hyper-parameter that measures the importance of $\mathcal{L}_{lcl}$ and can be adjusted.

\section{Experimental Settings}
\subsection{Datasets and Implementation Details}

To evaluate the effect of our framework, we carry out experiments on three datasets {\tt FewAsp(single)}, {\tt FewAsp(multi)}, and {\tt FewAsp} from \cite{DBLP:conf/acl/HuZGXGGCS20}, which share the same 100 aspects, with 64 aspects for training, 16 aspects for validation and 20 aspects for testing. It is notable that a sentence may belong to a single aspect or multiple aspects. {\tt FewAsp(single)}, {\tt FewAsp(multi)}, and {\tt FewAsp} are composed of single-aspect, multi-aspect, and both types of sentences, respectively. General information for three datasets is presented in Table \ref{datasets}.

In each dataset, we construct four FS-ACD tasks, where $N$ = 5, 10 and $K$ = 5, 10. And the number of query instances per class is 5. All the models are implemented by the Tensorflow framework with an NVIDIA Tesla V100 GPU. The hyperparameters and training details are given in \textbf{Appendix \ref{Implementation}}.




\subsection{Evaluation Metrics}



Following~\cite{DBLP:conf/acl/HuZGXGGCS20}, we use Macro-F1 and AUC scores as our evaluation metrics, and the thresholds in the 5-way setting and 10-way setting are set to \{0.3, 0.2\}, respectively. Besides, the paired $t$-test is conducted to test the significance of different approaches. Finally, we report the average performance and standard deviation over 5 runs, where the seeds are set to [5, 10, 15, 20, 25]. Our code and
datasets are available at \url{https://github.com/1429904852/LDF}.



\subsection{Compared Methods}

Following \cite{DBLP:conf/acl/HuZGXGGCS20}, we chose some frequently-used baselines: \emph{Matching Network} \cite{DBLP:conf/nips/VinyalsBLKW16}, \emph{Prototypical Network} \cite{DBLP:conf/nips/SnellSZ17}, \emph{Relation Network} \cite{DBLP:conf/cvpr/SungYZXTH18}, \emph{Graph Network} \cite{DBLP:conf/iclr/SatorrasE18}, \emph{IMP} \cite{DBLP:conf/icml/AllenSST19}, \emph{Proto-HATT} \cite{DBLP:conf/aaai/GaoH0S19} and \emph{Proto-AWATT} \cite{DBLP:conf/acl/HuZGXGGCS20}.



To verify the superiority of the LDF framework, we chose two dominant models with the best performance as the foundations of our work, i.e., \emph{Proto-HATT} and \emph{Proto-AWATT}. Finally, we integrate LDF into \emph{Proto-HATT} and \emph{Proto-AWATT} to obtain the model \emph{LDF-HATT} and \emph{LDF-AWATT}.

\begin{table}[t]
\centering
\scalebox{0.90}{
\begin{tabular}{l|c|c|c}
\toprule
\textbf{Dataset} & \textbf{\#cls.} & \textbf{\#inst./cls.} & \textbf{\#inst.}\\
\hline
FewAsp(single) & 100 & 200 & 20000 \\
FewAsp(multi) & 100 & 400 & 40000\\
FewAsp & 100 & 630 & 63000\\
\bottomrule
\end{tabular}}
\caption{Statistics of three datasets. \textbf{\#cls.} is the number of classes. \textbf{\#inst.} is the total number of instances. \textbf{\#inst./cls.} is the number of instances per class.}
 \label{datasets}%
\end{table}

\begin{table*}[t]
\centering
\scalebox{0.68}{
\begin{tabular}{l|cc|cc|cc|cc}
\toprule
\multirow{2}{*}{\textbf{Models}} & \multicolumn{2}{c|}{\textbf{5-way 5-shot}} & \multicolumn{2}{c|}{\textbf{5-way 10-shot}} & \multicolumn{2}{c|}{\textbf{10-way 5-shot}} & \multicolumn{2}{c}{\textbf{10-way 10-shot}}\\
& \textbf{F1} & \textbf{AUC} & \textbf{F1} & \textbf{AUC} & \textbf{F1} & \textbf{AUC} & \textbf{F1} & \textbf{AUC} \\
\midrule
\multicolumn{9}{c}{\textit{FewAsp}}\\
\midrule
Proto-HATT & 70.26 & 91.54 & 75.24 & 93.43 & 57.26 & 90.63 & 61.51 & 92.86 \\ 
\textbf{LDF-HATT} & \textbf{73.56}$^\dagger\pm$0.47 & \textbf{92.60}$^\dagger\pm$0.23 & \textbf{78.81}$^\dagger\pm$0.93 & \textbf{94.75}$^\dagger\pm$0.43 & \textbf{60.68}$^\dagger\pm$0.92 & \textbf{91.22}$\pm$0.53 & \textbf{67.13}$^\dagger\pm${0.94} &  \textbf{94.12}$^\dagger\pm${0.29} \\ 
\hspace{2.5em}$\Delta$ & {\color{black}{+}3.30} & {\color{black}{+}1.06} & {\color{black}{+}3.57} & {\color{black}{+}1.32} & {\color{black}{+}3.42} & {\color{black}{+}0.59} & {\color{black}{+}5.62} & {\color{black}{+}1.26} \\
\midrule 
Proto-AWATT & 75.37 & 93.35 & 80.16 & 95.28 & 65.65 & 92.06 & 69.70 & 93.42 \\ 
\textbf{LDF-AWATT} & \textbf{78.27}$^\dagger\pm${0.89} & \textbf{94.65}$^\dagger\pm${0.41} & \textbf{81.87}$^\dagger\pm${0.48} & \textbf{95.71}$\pm${0.26} & \textbf{67.13}$^\dagger\pm${0.41} & \textbf{92.74}$\pm${0.12} & \textbf{71.97}$^\dagger\pm${0.49} & \textbf{94.29}$\pm${0.25} \\
\hspace{2.5em}$\Delta$ & {\color{black}{+}2.90} & {\color{black}{+}1.30} & {\color{black}{+}1.71} & {\color{black}{+}0.43} & {\color{black}{+}1.48} & {\color{black}{+}0.68} & {\color{black}{+}2.27} & {\color{black}{+}0.87} \\
\midrule
\multicolumn{9}{c}{\textit{FewAsp(single)}}\\
\midrule
Proto-HATT & 83.33 & 96.45 & 86.71 & 97.62 & 73.42 & 95.71 & 77.65 & 97.00 \\ 
\textbf{LDF-HATT} & \textbf{84.41}$^\dagger\pm$0.46 & \textbf{97.06}$\pm$0.16 & \textbf{88.15}$^\dagger\pm$1.00 & \textbf{98.12}$\pm$0.31 & \textbf{76.27}$^\dagger\pm$1.08 & \textbf{96.38}$\pm$0.37 & \textbf{80.54}$^\dagger\pm$0.97 & \textbf{97.45}$\pm$0.14 \\ 
\hspace{2.5em}$\Delta$ & {\color{black}{+}1.08} & {\color{black}{+}0.61} & {\color{black}{+}1.44} & {\color{black}{+}0.50} & {\color{black}{+}2.85} & {\color{black}{+}0.67} & {\color{black}{+}2.89} & {\color{black}{+}0.45} \\
\midrule 
Proto-AWATT & 86.71 & 97.56 & 88.54 & 97.96 & 80.28 & 97.01 & 82.97 & 97.55 \\
\textbf{LDF-AWATT} & \textbf{88.16}$^\dagger\pm${0.62} & \textbf{98.29}$\pm$0.32 & \textbf{89.32}$\pm${0.92} & \textbf{98.38}$\pm${0.13} & \textbf{81.73}$^\dagger\pm${0.96} & \textbf{97.51}$\pm${0.33} & \textbf{84.20}$^\dagger\pm$0.21 & \textbf{97.96}$\pm$0.30 \\
\hspace{2.5em}$\Delta$ & {\color{black}{+}1.45} & {\color{black}{+}0.73} & {\color{black}{+}0.78} & {\color{black}{+}0.42} & {\color{black}{+}1.45} & {\color{black}{+}0.50} & {\color{black}{+}1.23} & {\color{black}{+}0.41} \\
\midrule
\multicolumn{9}{c}{\textit{FewAsp(multi)}}\\
\midrule
Proto-HATT & 69.15 & 91.10 & 73.91 & 93.03 & 55.34 & 90.44 & 60.21 & 92.38 \\ 
\textbf{LDF-HATT} & \textbf{72.13}$^\dagger\pm$0.79 & \textbf{92.19}$^\dagger\pm$0.33 & \textbf{76.52}$^\dagger\pm$0.74 & \textbf{93.68}$\pm$0.36 & \textbf{59.10}$^\dagger\pm$1.04 & \textbf{91.00}$\pm$0.51 & \textbf{65.31}$^\dagger\pm$0.57 & \textbf{92.99}$\pm$0.24 \\ 
\hspace{2.5em}$\Delta$ & {\color{black}{+}2.98} & {\color{black}{+}1.09} & {\color{black}{+}2.61} & {\color{black}{+}0.65} & {\color{black}{+}3.76} & {\color{black}{+}0.56} & {\color{black}{+}5.10} & {\color{black}{+}0.61} \\
\midrule
Proto-AWATT & 71.72 & 91.45 & 77.19 & 93.89 & 58.89 & 89.80 & 66.76 & 92.34 \\
\textbf{LDF-AWATT} & \textbf{73.38}$^\dagger\pm${0.73} & \textbf{92.62}$^\dagger\pm${0.32} & \textbf{78.81}$^\dagger\pm${0.19} & \textbf{94.34}$\pm${0.15} & \textbf{62.06}$^\dagger\pm${0.54} & \textbf{90.87}$^\dagger\pm${0.48} & \textbf{68.23}$^\dagger\pm${0.98} & \textbf{92.93}$\pm${0.44} \\
\hspace{2.5em}$\Delta$ & {\color{black}{+}1.66} & {\color{black}{+}1.17} & {\color{black}{+}1.62} & {\color{black}{+}0.44} & {\color{black}{+}3.17} & {\color{black}{+}1.07} & {\color{black}{+}1.47} & {\color{black}{+}0.59} \\
\bottomrule
\end{tabular}}
\caption{Test Macro-F1 and AUC score on the FewAsp, FewAsp(single), and FewAsp(multi) datasets (\%). The results of Proto-HATT and Proto-AWATT are retrieved from \cite{DBLP:conf/acl/HuZGXGGCS20}. We report the average performance and standard deviation over 5 runs, the thresholds in the 5-way setting and 10-way setting are set to \{0.3, 0.2\}. Best results are in bold. The marker $^\dagger$ refers to significant test p-value $<$ 0.05 when comparing with Proto-HATT and Proto-AWATT. $\Delta$ denotes the difference between the performance of Proto-HATT and LDF-HATT, as well as Proto-AWATT and LDF-AWATT. Due to space constraints, we report other baseline results in \textbf{Appendix \ref{main_result}}.}
\label{tab:mainresult}%
\end{table*}

\begin{table*}[t]
\centering
\scalebox{0.69}{
\begin{tabular}{ll|cc|cc|cc|cc}
\toprule
& \multirow{2}{*}{\textbf{Models}} & \multicolumn{2}{c|}{\textbf{5-way 5-shot}} & \multicolumn{2}{c|}{\textbf{5-way 10-shot}} & \multicolumn{2}{c|}{\textbf{10-way 5-shot}} & \multicolumn{2}{c}{\textbf{10-way 10-shot}}\\
& & \textbf{F1} & \textbf{AUC} & \textbf{F1} & \textbf{AUC} & \textbf{F1} & \textbf{AUC} & \textbf{F1} & \textbf{AUC} \\
\midrule
& Proto-AWATT & 75.37 & 93.35 & 80.16 & 95.28 & 65.65 & 92.06 & 69.70 & 93.42 \\ 
\midrule
& Proto-AWATT+LAS & 77.31$\pm$1.96 & 94.42$\pm$0.67 & 81.19$\pm$0.84 & 95.49$\pm$0.36 & 66.48$\pm$3.02 & 92.54$\pm$0.70 & 71.12$\pm$1.14 & 94.26$\pm$0.40 \\ 
& Proto-AWATT+LCL & 77.06$\pm$0.71 & 94.20$\pm$0.26 & 80.78$\pm$0.39 & 95.44$\pm$0.22 & 66.20$\pm$1.26 & 92.38$\pm$0.45 & 70.83$\pm$0.66 & 94.07$\pm$0.33 \\ 
& Proto-AWATT+SCL & 76.11$\pm$1.76 & 93.67$\pm$0.80 & 80.24$\pm$2.99 & 95.31$\pm$1.01 & 65.76$\pm$2.17 & 92.36$\pm$0.60 & 70.03$\pm$2.69 & 93.93$\pm$0.67 \\
\midrule
& \textbf{LDF-AWATT} & \textbf{78.27}$\pm${0.89} & \textbf{94.65}$\pm${0.41} & \textbf{81.87}$\pm${0.48} & \textbf{95.71}$\pm${0.26} & \textbf{67.13}$\pm${0.41} & \textbf{92.74}$\pm${0.12} & \textbf{71.97}$\pm${0.49} & \textbf{94.29}$\pm${0.25} \\
\bottomrule
\end{tabular}}
\caption{Ablation study over two main components on FewAsp dataset. The ablation results of FewAsp(single) and FewAsp(multi) datasets are included in \textbf{Appendix \ref{Ablation}}.}
 \label{ablation_study}%
\end{table*}

\section{Results and Discussion}
\subsection{Main Results}


The main experiment results are shown in Table~\ref{tab:mainresult}. From this table, we can see that: (1) \emph{LDF-HATT} and \emph{LDF-AWATT} consistently outperform their base models on three datasets. It is worth mentioning that \emph{LDF-HATT} at most obtains 5.62\% and 1.32\% improvements in Macro-F1 and AUC scores. In contrast, \emph{LDF-AWATT} outperforms \emph{Proto-AWATT} by 3.17\% and 1.30\% at most. These results reveal that our framework has good compatibility; (2) It is a fact that the Macro-F1 of \emph{LDF-AWATT} is improved by about 2\% in most settings, while that of \emph{LDF-HATT} is improved by about 3\% on average. This is consistent with our expectations since the original \emph{Proto-AWATT} has a more powerful performance; (3) \emph{LDF-HATT} and \emph{LDF-AWATT} perform better on the {\tt FewAsp(multi)} dataset than on the {\tt FewAsp(single)} dataset. A possible reason is that each class in the {\tt FewAsp(multi)} dataset contains more instances, which allows \emph{LDF-HATT} and \emph{LDF-AWATT} to generate a more accurate prototype in multi-label classification.



\subsection{Ablation Study}

Without loss of generality, we choose \emph{LDF-AWATT} model for the ablation study to investigate the effects of different components in LDF\footnote{Due to space limitations, we report the ablation results of \emph{LDF-HATT} in \textbf{Appendix \ref{Ablation}}.}.


\paragraph{Effect of Label-Driven Denoising Framework.}We study the two main components of LDF: Label-guided Attention Strategy (LAS) and Label-weighted Contrastive Loss (LCL). Based on the results in Table~\ref{ablation_study}, we can make a couple of observations: (1) Compared to the base model \emph{Proto-AWATT}, \emph{Proto-AWATT+LAS} achieves competitive performance on three datasets, which validates the rationality of exploiting label information to generate a better prototype for each class; (2) After integrating LCL into \emph{Proto-AWATT+LAS}, \emph{LDF-AWATT} achieve the state-of-the-art performance, which demonstrates that LCL is beneficial to distinguish similar prototypes; (3) LAS is more effective than LCL. A possible reason is that the attention mechanism is the core factor in producing the prototype. Hence, it contributes more to our framework.

\paragraph{Analysis of Label in Contrastive Loss.} We compare Lable-weighted Contrastive Loss (LCL) with the Supervised Contrastive Loss (SCL) to see the contribution of label. It can be seen from Table \ref{ablation_study} that: (1) \emph{Proto-AWATT+SCL} performs slightly better than \emph{Proto-AWATT} on {\tt FewAsp} dataset, but their results are much lower than \emph{Proto-AWATT+LCL}. These results further highlight the effectiveness of LCL; (2) After integrating inter-class relationships into \emph{Proto-AWATT+SCL}, \emph{Proto-AWATT+LCL} achieve better performance, which indicates that the inter-class relationships play a crucial role in distinguishing similar prototypes.

\begin{figure*}[t]
\centering
\subfigure[FewAsp]{
\begin{minipage}[t]{0.33\linewidth}
\centering
\includegraphics[width=2.0in]{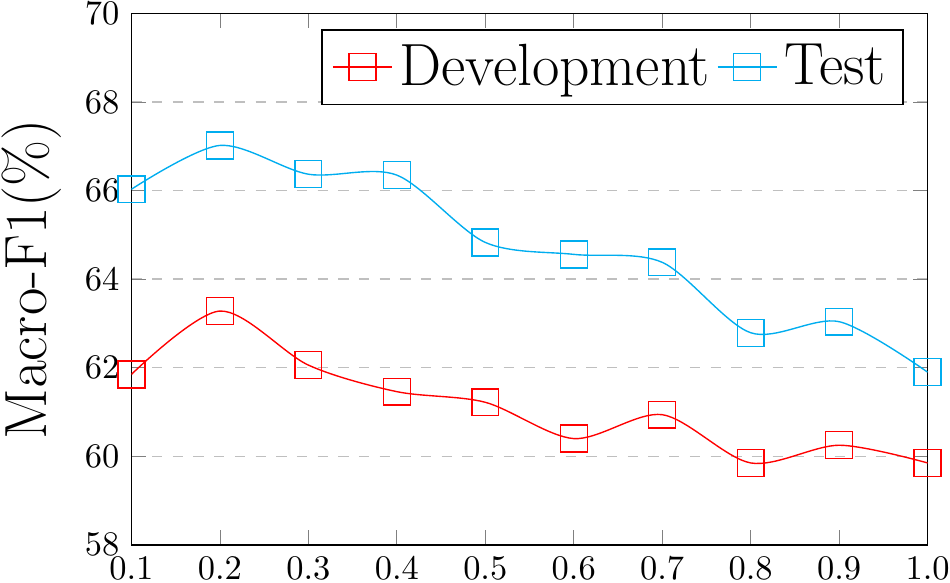}
\label{hyper_1}
\end{minipage}%
}%
\subfigure[FewAsp(single)]{
\begin{minipage}[t]{0.33\linewidth}
\centering
\includegraphics[width=2.0in]{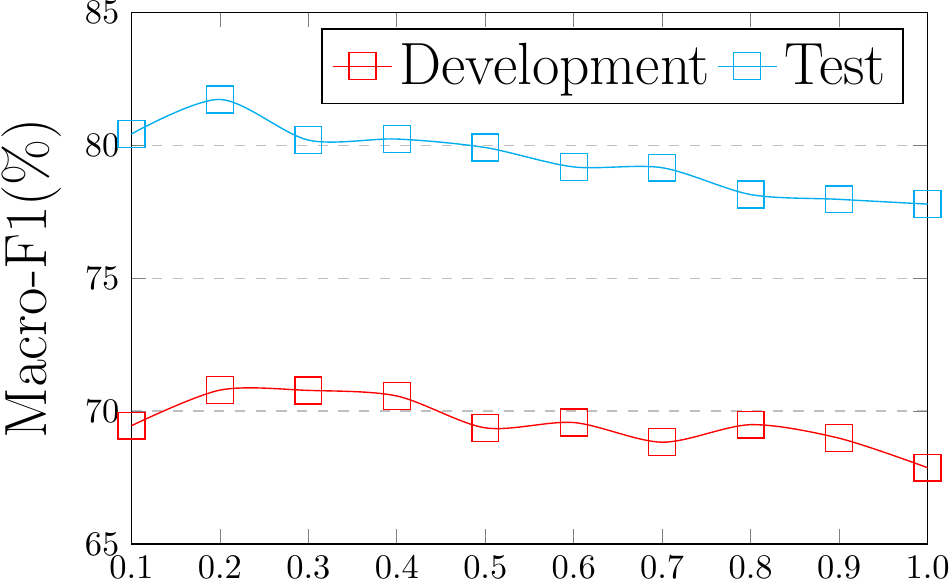}
\label{hyper_2}
\end{minipage}%
}%
\subfigure[FewAsp(multi)]{
\begin{minipage}[t]{0.33\linewidth}
\centering
\includegraphics[width=2.0in]{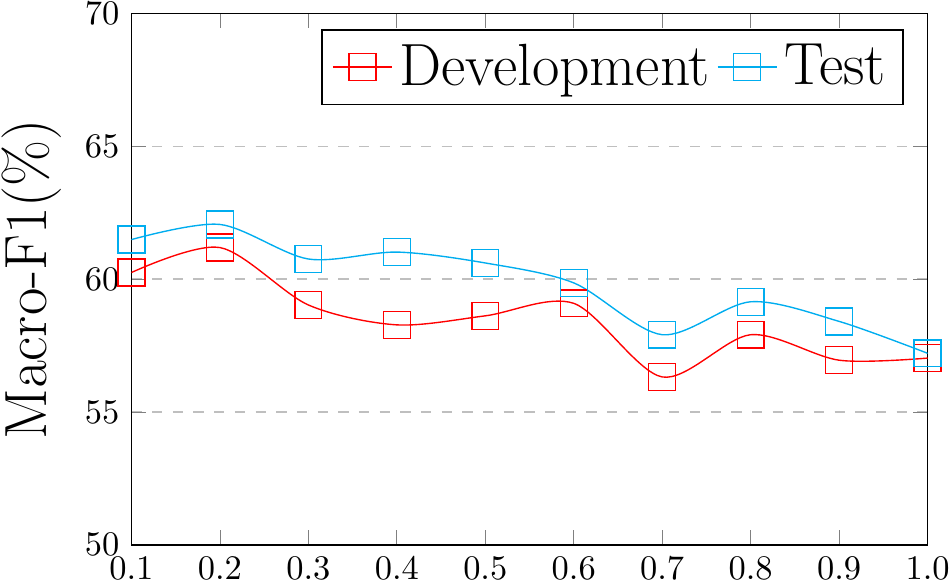}
\label{hyper_3}
\end{minipage}
}%
\centering
\caption{Effect of $\lambda$ in the 10-way 5-shot setting on three dataset.}
\label{parameter}
\end{figure*}

\begin{table}[t]
\centering
\scalebox{0.65}{
\begin{tabular}{l|cc|cc}
\toprule
\multirow{2}{*}{Models} & \multicolumn{2}{c|}{\textbf{GloVe + CNN	}} & \multicolumn{2}{c}{\textbf{BERT}} \\
& \textbf{F1} & \textbf{AUC} & \textbf{F1} & \textbf{AUC} \\
\midrule
Proto-HATT$^\clubsuit$ & 57.26 & 90.63 & 57.33 & 89.70 \\
LDF-HATT & 60.68$\pm$0.92 & 91.22$\pm$0.53 & 63.72$\pm$0.27 & 91.99$\pm$0.12 \\
\midrule
Proto-AWATT$^\clubsuit$ & 65.65 & 92.06 & 70.09 & 94.59 \\
LDF-AWATT & 67.13$\pm$0.41 & 92.74$\pm$0.12 & 72.76$\pm$0.29 & 95.31$\pm$0.19 \\
\bottomrule
\end{tabular}}
\caption{The effect of different encoders in the 10-way 5-shot scenario on FewAsp dataset. The results with symbol $^\clubsuit$ are retrieved from~\cite{DBLP:conf/acl/HuZGXGGCS20}.}
\label{tab:bert}%
\end{table}


\subsection{Discussion}

\paragraph{Effect of Encoder.}We also conduct experiments (shown in Table~\ref{tab:bert}) using the pre-trained BERT model \cite{DBLP:conf/naacl/DevlinCLT19}. Concretely, we replace the Glove+CNN encoder with BERT and keep the other components the same as our original model. It's clear that \emph{LDF-AWATT} and \emph{LDF-HATT} perform remarkably well than the base model \emph{Proto-AWATT} and \emph{Proto-HATT} on all encoders, which proves that our framework has good scalability.



\begin{table}[t]
\centering
    \scalebox{0.68}{
        \begin{tabular}{l|cc}
        \toprule
        \multirow{2}{*}{Models} & \multicolumn{2}{c}{\textbf{10-way 5-shot}} \\
        & \textbf{F1} & \textbf{AUC}\\
        \midrule
        Proto-AWATT & 65.65 & 92.06 \\
        Proto-AWATT (LSW) & 57.84$\pm$0.49 & 90.85$\pm$0.22\\
        \bottomrule
        \end{tabular}
    }
\caption{The effect of label similarity weight $\alpha$ in the 10-way 5-shot scenario on FewAsp dataset.}
\label{tab:alpha}%
\end{table}


\paragraph{Effect of Label Similarity Weight $\alpha$.}
To illustrate the role of the similarity weight $\alpha$, we directly replace the attention weight $\beta$ in Equation \ref{beta} with the similarity weight $\alpha$ in Equation \ref{alpha}, and name this method as \emph{Proto-AWATT(LSW)}. From the results in Table~\ref{tab:alpha}, we can see that the performance of \emph{Proto-AWATT(LSW)} is far inferior to \emph{Proto-AWATT}, which implies that the similarity weight only plays a supporting role to the attention weight, and cannot be treated independently for the FS-ACD task.


\paragraph{Effect of hyper-parameter $\lambda$.}We tune the hyper-parameter $\lambda$ on the development set of each dataset, and then evaluate the performance of \emph{LDF-AWATT} on the test set. Specifically, we conduct experiments for values set at 0.1 intervals in the range (0, 1). Figure \ref{parameter} shows the performance of \emph{LDF-AWATT} with different $\lambda$ on three dataset. Actually, as $\lambda$ increases, the performance of \emph{LDF-AWATT} has an initial upward trend, and then flattens out or begins to fall. In the upward part, the Label-weighted Contrastive Loss (LCL) is useful guidance to help the \emph{LDF-AWATT} distinguish similar prototypes more accurately, thus improving the performance. However, once the weight $\lambda$ exceeds 0.2, the LCL begins to dominate and performs poorly. The reason behind this may be that the bigger $\lambda$ has a negative effect on the MSE loss of the model. Therefore, we set $\lambda$ to be 0.2 on three datasets. In addition, we find that the best results of the development set and test set are basically consistent, which indicates that our framework has good robustness.

\begin{figure}[t]
\centering
\includegraphics[width=2.9in]{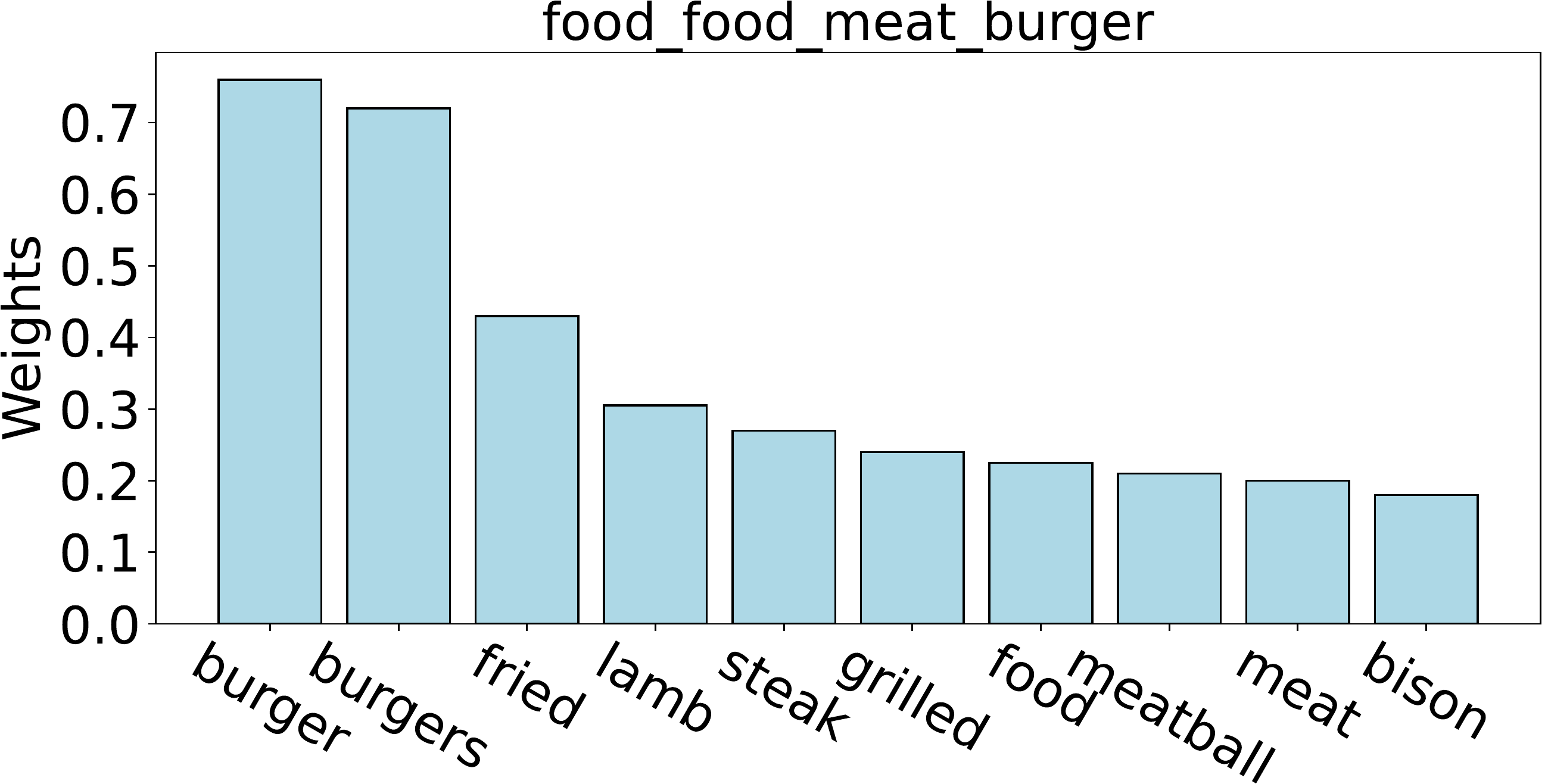}
\caption{Visualize the top-10 words for the prototype of aspect category \emph{food\_food\_meat\_burger} based on the attention weights of \emph{Proto-AWATT+LAS}.}
\label{figure_label_a}
\end{figure}


\begin{figure*}[t]
\centering
\subfigure[Proto-AWATT]{
\begin{minipage}[t]{0.32\linewidth}
\centering
\includegraphics[width=1.9in]{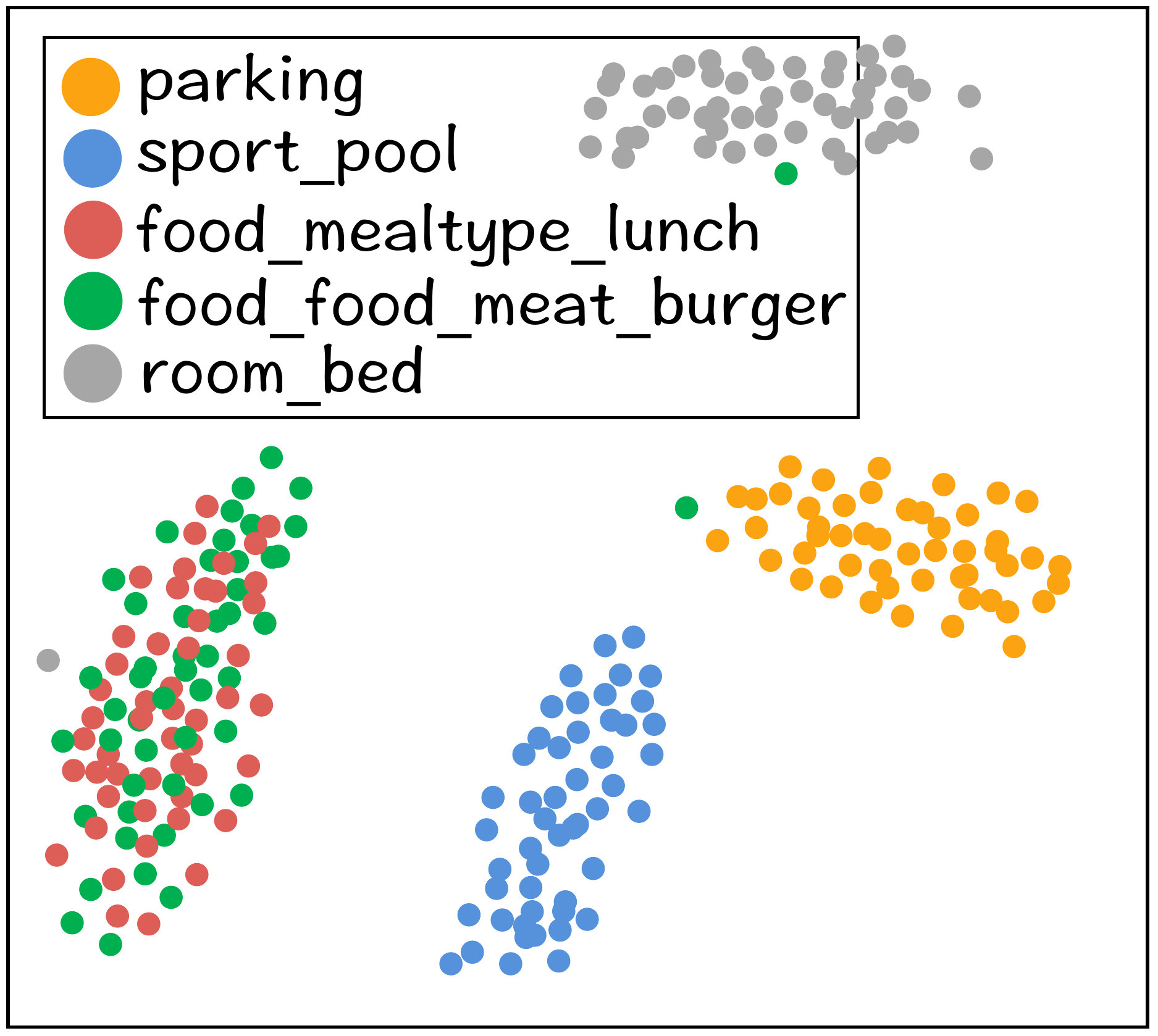}
\end{minipage}%
}
\subfigure[Proto-AWATT+LAS]{
\begin{minipage}[t]{0.32\linewidth}
\centering
\includegraphics[width=1.9in]{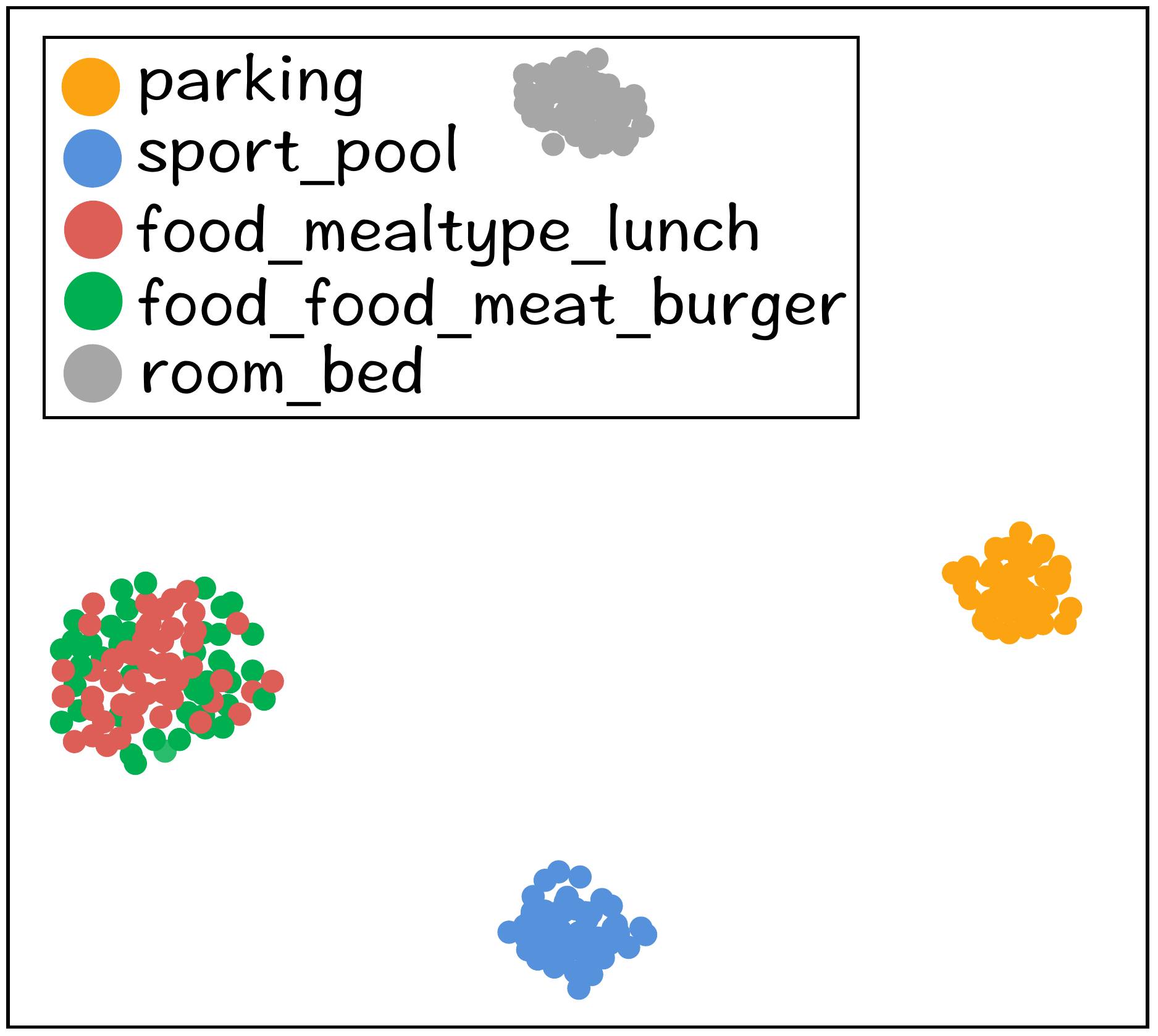}
\end{minipage}%
}
\subfigure[LDF-AWATT]{
\begin{minipage}[t]{0.32\linewidth}
\centering
\includegraphics[width=1.9in]{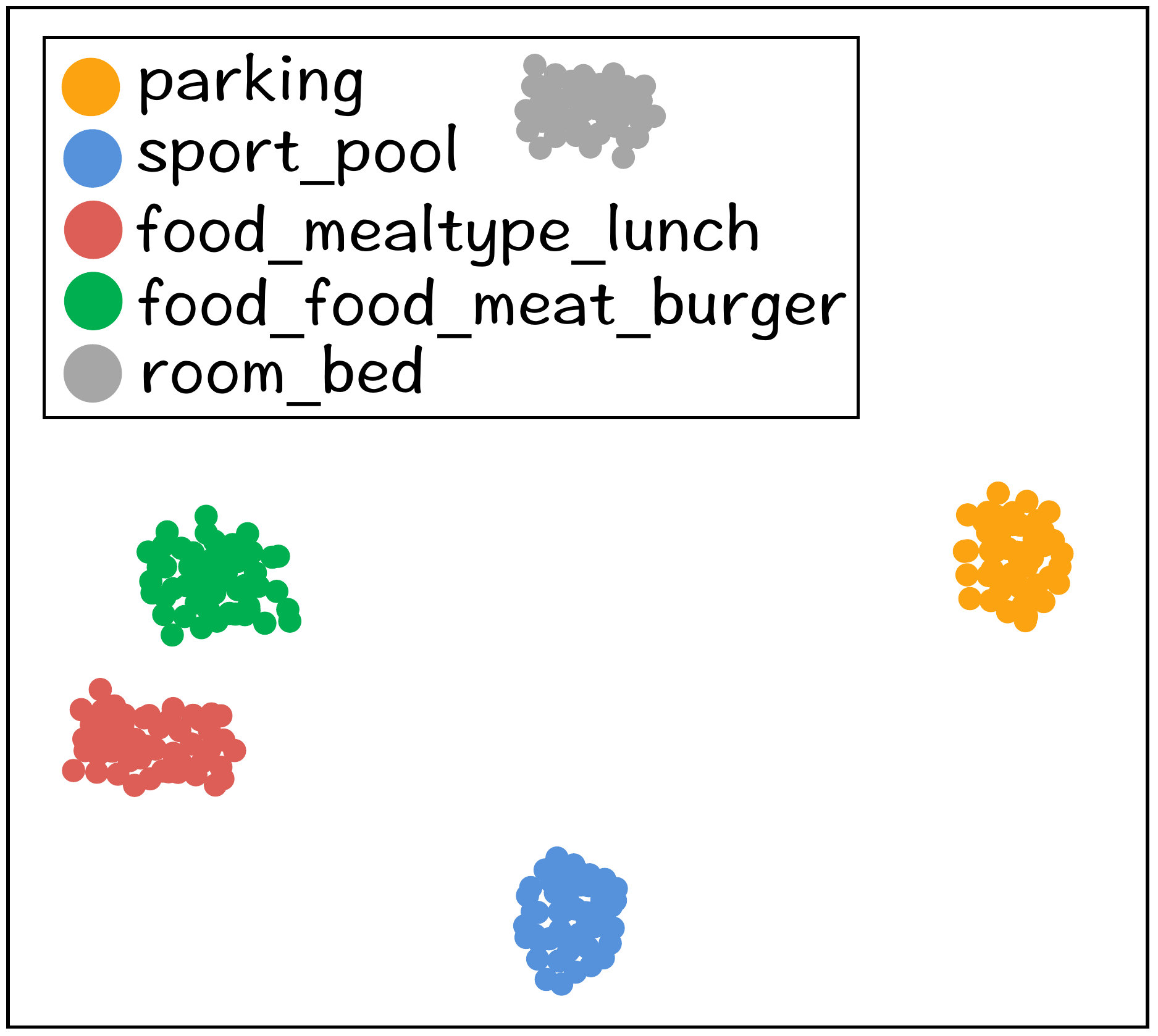}
\end{minipage}%
}%
\centering
\caption{Visualization of prototype representations for Proto-AWATT, Proto-AWATT+LAS and LDF-AWATT.}
\label{fig:vis1}
\end{figure*}

\subsection{Case Study}

To better understand the advantage of our framework, we select some samples from {\tt FewAsp} dataset for a case study. Specifically, we randomly sample 5 classes and then sample 50 times of 5-way 5-shot meta-tasks for the five classes. Finally for each class, we obtain 50 prototype vectors\footnote{visualized by t-SNE \cite{van2008visualizing}.}.


\paragraph{Proto-AWATT vs. Proto-AWATT+LAS.}As shown in Figure \ref{fig:vis1}(a) and Figure \ref{fig:vis1}(b), we can see that the prototype representation for each class learned by \emph{Proto-AWATT+LAS} are obviously more concentrated than those by \emph{Proto-AWATT}. Besides, in contrast to \emph{Proto-AWATT}, \emph{Proto-AWATT+LAS} can focus on class-relevant words better (shown in Figure \ref{figure_a} and Figure \ref{figure_label_a}). 
These observations suggest that \emph{Proto-AWATT+LAS} can indeed generate a more accurate prototype for each class.





\paragraph{Proto-AWATT+LAS vs. LDF-AWATT.}As depicted in Figure \ref{fig:vis1}(b) and \ref{fig:vis1}(c), after incorporating LCL into \emph{Proto-AWATT+LAS}, the prototype representation of ``{\it food\_mealtype\_lunch}'' and ``{\it food\_food\_meat\_burger}'' learned by \emph{LDF-AWATT} are more separable than those by \emph{Proto-AWATT+LAS}. This reveals that LCL can indeed distinguish similar prototypes.

\subsection{Error Analysis}

We present the error analysis in the \textbf{Appendix \ref{error}}.

\section{Related Work}

\paragraph{Aspect Category Detection.}
Previous works formulate ACD in a data-driven scenario, and can be generally divided into two kinds: one is unsupervised approach, which detects aspect categories by exploiting semantic association \cite{DBLP:conf/iccpol/SuXWSY06} or co-occurrence frequency \cite{DBLP:conf/cicling/HaiCK11,DBLP:journals/tcyb/SchoutenWFD18}; the other is supervised approach, which uses hand-crafted features \cite{DBLP:conf/semeval/KiritchenkoZCM14}, learns useful representations automatically \cite{DBLP:conf/aaai/ZhouWX15}, adopts a multi-task learning strategy \cite{DBLP:conf/emnlp/HuZZCSCS19}, or utilizes a topic-attention model \cite{DBLP:journals/corr/abs-1901-01183} to address the ACD task. However, the above methods heavily rely on large-scale training data, which is time-consuming to annotate. 

\paragraph{Multi-Label Few-Shot Learning.}

In comparison with single-label FSL, multi-label FSL is more difficult and less explored, as it aims to identify multiple labels for an instance. ~\citeauthor{DBLP:conf/emnlp/RiosK18}~\shortcite{DBLP:conf/emnlp/RiosK18} propose few-shot learning methods for multi-label text classification over a structured label space. Further research on multi-label FSL are developed on image synthesis ~\cite{DBLP:conf/cvpr/AlfassyKASHFGB19}, signal processing ~\cite{DBLP:conf/mmsp/ChengCY19}, and intent detection ~\cite{DBLP:conf/aaai/HouLWCL21}. Recently, ~\citeauthor{DBLP:conf/acl/HuZGXGGCS20}~\shortcite{DBLP:conf/acl/HuZGXGGCS20} formalize aspect category detection in a multi-label few-shot scenario to alleviate the dependency on large-scale labeled data. However, ~\citeauthor{DBLP:conf/acl/HuZGXGGCS20}~\shortcite{DBLP:conf/acl/HuZGXGGCS20} ignore the label information of each class, which is crucial for generating a representative prototype in the FS-ACD task.

\paragraph{Contrastive Learning.}
Contrastive Learning is a representation learning technique and has proven its effectiveness in the field of natural language processing \cite{DBLP:conf/iclr/GunelDCS21,DBLP:conf/acl/KimYL20,DBLP:conf/emnlp/YeKO21}. With the help of label information, \citeauthor{DBLP:conf/nips/KhoslaTWSTIMLK20} \shortcite{DBLP:conf/nips/KhoslaTWSTIMLK20} propose supervised contrastive learning, which aims to improve the quality of learnt representations in a supervised setting. Different from their work, we do not treat label information equally and propose a label-weighted contrastive loss to distinguish similar prototypes.

\section{Conclusion}
In this paper, we propose a novel Label-Driven Denoising Framework (LDF) to alleviate the noise problems for the FS-ACD task. Specifically, we design two reasonable components: Label-guided Attention Strategy and Label-weighted Contrastive Loss, which aim to produce a  better prototype for each class and distinguish similar prototypes. Results from numerous experiments indicate that our framework LDF achieves better performance than other state-of-the-art methods.

\section*{Limitations}

We consider two major limitations in the FS-ACD task that need to be addressed in current research and related fields: (1) Existing studies for few-shot learning (FSL) require both the training and testing data have the same number of classes (denoted as $N$-way) and the same number of instances in each class (denoted as $K$-shot) in the support set. However, little investigation has been done towards inconsistent classes and inconsistent instances per class during training and testing. As far as we know, inconsistent FSL is more realistic and meaningful, which may be extremely helpful in low-resource scenarios; (2) The FS-ACD models usually give incorrect predictions when a sentence belongs to more than four aspect categories. A possible reason is that these sentences account for a small proportion of the dataset. Thus, it is also important to find effective methods to tackle the long-tail problem in multi-label classification. In general, the above limitations are of practical meaning and need us to do further research and exploration.

\section*{Acknowledgements}
We would like to thank the anonymous reviewers for their constructive comments. This work was supported by the National Natural Science Foundation of China (No. 61936012 and 61976114).

\bibliography{custom}
\bibliographystyle{acl_natbib}

\appendix
\onecolumn
\clearpage



\section{Appendices}

\subsection{Implementation Details}\label{Implementation}
\paragraph{Hyperparameters.}

We initialize word embedding with 50-dimension Glove vectors. All other parameters are initialized by sampling from a normal distribution $\mathcal{N}(0, 0.1)$. The hyper-parameter $\lambda$ is set to 0.2 on three datasets. The dimension of the hidden state is set to 50. The convolutional window size is set as 3. The optimizer is Adam with a learning rate 10$^{-3}$ and the temperature $\tau$ is set to 0.1. In each dataset, we construct four FS-ACD tasks, where N = 5, 10 and K = 5, 10. And the number of query instances per class is 5. For example, in a 5-way 10-shot meta-task, there are 5 $\times$ 10 = 50 instances in the support set and 5 $\times$ 5 = 25 instances in the query set.

\paragraph{Training Details.}


During training, we train each model for a fixed 30 epochs, and then select the model with the best AUC score on the development set. Finally, we evaluate its performance on the test set. In every epoch, we randomly sample 800 meta-tasks for training. The number of meta-tasks during validation and testing are both
set as 600. Besides, we employ an early stop strategy if the AUC score of the validation set is not improved in 3 epochs. For all baselines and our model, we report the average testing results from 5 runs, where the seeds are set to [5, 10, 15, 20, 25]. All the models are implemented by the Tensorflow framework with an NVIDIA Tesla V100 GPU.

\begin{table*}[htbp]
\centering
\scalebox{0.70}{
\begin{tabular}{l|cc|cc|cc|cc}
\toprule
\multirow{2}{*}{\textbf{Models}} & \multicolumn{2}{c|}{\textbf{5-way 5-shot}} & \multicolumn{2}{c|}{\textbf{5-way 10-shot}} & \multicolumn{2}{c|}{\textbf{10-way 5-shot}} & \multicolumn{2}{c}{\textbf{10-way 10-shot}}\\
& \textbf{F1} & \textbf{AUC} & \textbf{F1} & \textbf{AUC} & \textbf{F1} & \textbf{AUC} & \textbf{F1} & \textbf{AUC} \\
\midrule
\multicolumn{9}{c}{\textit{FewAsp}}\\
\midrule
Relation Network & 59.52 & 85.56 & 62.78 & 86.98 & 45.62 & 84.94 & 44.70 & 83.77 \\
Matching Network & 67.14 & 90.76 & 70.09 & 92.39 & 51.27 & 88.44 & 54.61 & 89.90 \\
Graph Network & 61.49 & 89.48 & 69.89 & 92.35 & 47.91 & 87.35 & 56.06 & 90.19 \\
Prototypical Network & 66.96 & 88.88 & 73.27 & 91.77 & 52.06 & 87.35 & 59.03 & 90.13 \\
IMP & 68.96 & 89.95 & 74.13 & 92.30 & 54.14 & 88.50 & 59.84 & 90.81 \\ 
\midrule
Proto-HATT & 70.26 & 91.54 & 75.24 & 93.43 & 57.26 & 90.63 & 61.51 & 92.86 \\ 
\textbf{LDF-HATT} & \textbf{73.56}$^\dagger\pm$0.47 & \textbf{92.60}$^\dagger\pm$0.23 & \textbf{78.81}$^\dagger\pm$0.93 & \textbf{94.75}$^\dagger\pm$0.43 & \textbf{60.68}$^\dagger\pm$0.92 & \textbf{91.22}$\pm$0.53 & \textbf{67.13}$^\dagger\pm${0.94} &  \textbf{94.12}$^\dagger\pm${0.29} \\ 
\midrule 
Proto-AWATT & 75.37 & 93.35 & 80.16 & 95.28 & 65.65 & 92.06 & 69.70 & 93.42 \\ 
\textbf{LDF-AWATT} & \textbf{78.27}$^\dagger\pm${0.89} & \textbf{94.65}$^\dagger\pm${0.41} & \textbf{81.87}$^\dagger\pm${0.48} & \textbf{95.71}$\pm${0.26} & \textbf{67.13}$^\dagger\pm${0.41} & \textbf{92.74}$\pm${0.12} & \textbf{71.97}$^\dagger\pm${0.49} & \textbf{94.29}$\pm${0.25} \\
\midrule
\multicolumn{9}{c}{\textit{FewAsp(single)}}\\
\midrule
Relation Network & 75.79 & 93.31 & 72.02 & 90.86 & 63.78 & 91.81 & 61.15 & 90.54 \\
Matching Network & 81.89 & 97.05 & 84.62 & 97.49 & 70.95 & 96.30 & 73.28 & 96.72 \\
Graph Network & 81.45 & 96.54 & 85.04 & 97.46 & 70.75 & 95.45 & 77.84 & 96.97 \\
Prototypical Network & 83.30 & 96.49 & 86.29 & 97.53 & 74.23 & 95.97 & 76.83 & 96.71 \\
IMP & 83.69 & 96.65 & 86.14 & 97.47 & 73.80 & 96.00 & 77.09 & 96.91 \\
\midrule 
Proto-HATT & 83.33 & 96.45 & 86.71 & 97.62 & 73.42 & 95.71 & 77.65 & 97.00 \\ 
\textbf{LDF-HATT} & \textbf{84.41}$^\dagger\pm$0.46 & \textbf{97.06}$\pm$0.16 & \textbf{88.15}$^\dagger\pm$1.00 & \textbf{98.12}$\pm$0.31 & \textbf{76.27}$^\dagger\pm$1.08 & \textbf{96.38}$\pm$0.37 & \textbf{80.54}$^\dagger\pm$0.97 & \textbf{97.45}$\pm$0.14 \\ 
\midrule 
Proto-AWATT & 86.71 & 97.56 & 88.54 & 97.96 & 80.28 & 97.01 & 82.97 & 97.55 \\
\textbf{LDF-AWATT} & \textbf{88.16}$^\dagger\pm${0.62} & \textbf{98.29}$\pm$0.32 & \textbf{89.32}$\pm${0.92} & \textbf{98.38}$\pm${0.13} & \textbf{81.73}$^\dagger\pm${0.96} & \textbf{97.51}$\pm${0.33} & \textbf{84.20}$^\dagger\pm$0.21 & \textbf{97.96}$\pm$0.30 \\
\midrule
\multicolumn{9}{c}{\textit{FewAsp(multi)}}\\
\midrule
Relation Network & 58.38 & 84.91 & 61.37 & 86.21 & 43.71 & 84.22 & 44.85 & 84.72 \\
Matching Network & 65.70 & 89.54 & 69.02 & 91.38 & 50.86 & 88.28 & 54.42 & 89.94 \\
Graph Network & 59.25 & 87.97 & 64.63 & 90.45 & 45.42 & 86.05 & 48.49 & 88.44 \\
Prototypical Network & 67.88 & 89.67 & 72.32 & 91.60 & 52.72 & 88.01 & 58.92 & 90.68 \\
IMP & 68.86 & 90.12 & 73.51 & 92.29 & 53.96 & 88.71 & 59.86 & 91.10\\
\midrule
Proto-HATT & 69.15 & 91.10 & 73.91 & 93.03 & 55.34 & 90.44 & 60.21 & 92.38 \\ 
\textbf{LDF-HATT} & \textbf{72.13}$^\dagger\pm$0.79 & \textbf{92.19}$^\dagger\pm$0.33 & \textbf{76.52}$^\dagger\pm$0.74 & \textbf{93.68}$\pm$0.36 & \textbf{59.10}$^\dagger\pm$1.04 & \textbf{91.00}$\pm$0.51 & \textbf{65.31}$^\dagger\pm$0.57 & \textbf{92.99}$\pm$0.24 \\ 
\midrule
Proto-AWATT & 71.72 & 91.45 & 77.19 & 93.89 & 58.89 & 89.80 & 66.76 & 92.34 \\
\textbf{LDF-AWATT} & \textbf{73.38}$^\dagger\pm${0.73} & \textbf{92.62}$^\dagger\pm${0.32} & \textbf{78.81}$^\dagger\pm${0.19} & \textbf{94.34}$\pm${0.15} & \textbf{62.06}$^\dagger\pm${0.54} & \textbf{90.87}$^\dagger\pm${0.48} & \textbf{68.23}$^\dagger\pm${0.98} & \textbf{92.93}$\pm${0.44} \\
\bottomrule
\end{tabular}}
\caption{Test Macro-F1 and AUC score on the FewAsp, FewAsp(single), and FewAsp(multi) datasets (\%) . The baseline results are retrieved from \cite{DBLP:conf/acl/HuZGXGGCS20}. We report the average performance and standard deviation over 5 runs, the thresholds in the 5-way setting and 10-way setting are set to \{0.3, 0.2\}. Best results are in bold. The marker $^\dagger$ refers to significant test p-value $<$ 0.05 when comparing with Proto-HATT and Proto-AWATT.}
\label{ap:tab:mainresult}%
\end{table*}

\subsection{Main Result}\label{main_result}

As shown in Table \ref{ap:tab:mainresult}, we list all the frequently-used baselines and our enhanced version. It is clear that \emph{Proto-HATT} and \emph{Proto-AWATT} consistently outperform other baselines, thus we chose them as the foundation of our work. Besides, we observe that our framework achieves better performance compared to all the baselines.

\subsection{Ablation Study}\label{Ablation}

In Table \ref{ap:ablation_study} and \ref{ap:ablation_study_a}, we present the ablation results of \emph{LDF-HATT} and \emph{LDF-AWATT} in details.

\begin{table*}[htbp]
\centering
\scalebox{0.75}{
\begin{tabular}{l|cc|cc|cc|cc}
\toprule
\multirow{2}{*}{\textbf{Models}} & \multicolumn{2}{c|}{\textbf{5-way 5-shot}} & \multicolumn{2}{c|}{\textbf{5-way 10-shot}} & \multicolumn{2}{c|}{\textbf{10-way 5-shot}} & \multicolumn{2}{c}{\textbf{10-way 10-shot}}\\
& \textbf{F1} & \textbf{AUC} & \textbf{F1} & \textbf{AUC} & \textbf{F1} & \textbf{AUC} & \textbf{F1} & \textbf{AUC} \\
\midrule
\multicolumn{9}{c}{\textit{FewAsp}}\\
\midrule
Proto-HATT & 70.26 & 91.54 & 75.24 & 93.43 & 57.26 & 90.63 & 61.51 & 92.86 \\ 
\midrule
Proto-HATT+LAS & 73.02$\pm$0.69 & 92.56$\pm$0.37 & 78.09$\pm$0.90 & 94.16$\pm$0.29 & 60.10$\pm$1.24 & 90.95$\pm$0.62 & 65.95$\pm${1.39} & 93.88$\pm${0.52} \\ 
Proto-HATT+LCL & 72.77$\pm${0.75} & 92.43$\pm$0.88 & 77.42$\pm$0.46 & 94.04$\pm$0.20 & 59.85$\pm${1.37} & 90.89$\pm$0.74 & 65.05$\pm${0.70} & 93.48$\pm${0.14} \\ 
Proto-HATT+SCL & 71.56$\pm$1.07 & 91.85$\pm$0.53 & 76.20$\pm$0.42 & 93.63$\pm$0.18 & 58.51$\pm$0.56 & 90.85$\pm$0.45 & 62.86$\pm$0.68 & 93.27$\pm$0.28 \\
\midrule
\multicolumn{9}{c}{\textit{FewAsp(single)}}\\
\midrule
Proto-HATT & 83.33 & 96.45 & 86.71 & 97.62 & 73.42 & 95.71 & 77.65 & 97.00 \\ 
\midrule
Proto-HATT+LAS & 83.96$\pm$0.23 & 96.92$\pm${0.27} & 87.80$\pm$1.02 & 98.12$\pm$0.30 & 75.82$\pm$0.49 & 96.15$\pm${0.14} & 79.90$\pm$1.04 & 97.24$\pm$0.24 \\ 
Proto-HATT+LCL & 83.89$\pm$0.81 & 96.88$\pm$0.27 & 87.54$\pm${1.19} & 97.86$\pm$0.36 & 75.48$\pm$0.61 & 96.12$\pm$0.14 & 79.66$\pm${1.05} & 97.16$\pm$0.55 \\ 
Proto-HATT+SCL & 83.35$\pm$0.70 & 96.80$\pm$0.23 & 86.96$\pm${0.79} & 97.67$\pm$0.44 & 74.60$\pm$0.47 & 96.00$\pm$0.16 & 78.55$\pm$1.06 & 97.16$\pm$0.18 \\
\midrule
\multicolumn{9}{c}{\textit{FewAsp(multi)}}\\
\midrule
Proto-HATT & 69.15 & 91.10 & 73.91 & 93.03 & 55.34 & 90.44 & 60.21 & 92.38 \\ 
\midrule
Proto-HATT+LAS & 71.44$\pm$0.54 & 91.74$\pm$0.25 & 76.17$\pm${1.14} & 93.50$\pm$0.45 & 58.50$\pm${0.65} & 90.72$\pm$0.48 & 64.76$\pm${0.83} & 92.62$\pm$0.47 \\ 
Proto-HATT+LCL & 71.15$\pm$0.38 & 91.59$\pm$0.10 & 75.86$\pm${0.49} & 93.47$\pm$0.58 & 57.90$\pm${0.96} & 90.50$\pm${0.45} & 64.65$\pm${0.72} & 92.57$\pm$0.47 \\ 
Proto-HATT+SCL & 70.37$\pm$0.39 & 91.41$\pm$0.15 & 74.82$\pm$0.88 & 93.32$\pm$0.50 & 56.72$\pm$0.91 & 90.49$\pm$0.40 & 62.77$\pm$0.96 & 92.40$\pm$0.37 \\
\bottomrule
\end{tabular}}
\caption{Ablation study over two main components of LDF-HATT. Besides, we also report the ablation result of Proto-HATT+LCL. We report the average performance and standard deviation over 5 runs.}
 \label{ap:ablation_study}%
\end{table*}

\begin{table*}[htbp]
\centering
\scalebox{0.75}{
\begin{tabular}{l|cc|cc|cc|cc}
\toprule
\multirow{2}{*}{\textbf{Models}} & \multicolumn{2}{c|}{\textbf{5-way 5-shot}} & \multicolumn{2}{c|}{\textbf{5-way 10-shot}} & \multicolumn{2}{c|}{\textbf{10-way 5-shot}} & \multicolumn{2}{c}{\textbf{10-way 10-shot}}\\
& \textbf{F1} & \textbf{AUC} & \textbf{F1} & \textbf{AUC} & \textbf{F1} & \textbf{AUC} & \textbf{F1} & \textbf{AUC} \\
\midrule
\multicolumn{9}{c}{\textit{FewAsp}}\\
\midrule
Proto-AWATT & 75.37 & 93.35 & 80.16 & 95.28 & 65.65 & 92.06 & 69.70 & 93.42 \\ 
\midrule
Proto-AWATT+LAS & 77.31$\pm${0.96} & 94.42$\pm${0.36} & 81.19$\pm$0.84 & 95.49$\pm$0.36 & 66.48$\pm${1.32} & 92.54$\pm${0.64} & 71.12$\pm$1.14 & 94.26$\pm$0.40 \\ 
Proto-AWATT+LCL & 77.06$\pm$0.71 & 94.20$\pm$0.26 & 80.78$\pm$0.39 & 95.44$\pm$0.22 & 66.20$\pm${1.02} & 92.38$\pm$0.45 & 70.83$\pm$0.66 & 94.07$\pm$0.33 \\ 
Proto-AWATT+SCL & 76.11$\pm${0.92} & 93.67$\pm${0.55} & 80.24$\pm${0.60} & 95.31$\pm${0.25} & 65.76$\pm${0.97} & 92.36$\pm${0.33} & 70.03$\pm${1.15} & 93.93$\pm${0.21} \\
\midrule
\multicolumn{9}{c}{\textit{FewAsp(single)}}\\
\midrule
Proto-AWATT & 86.71 & 97.56 & 88.54 & 97.96 & 80.28 & 97.01 & 82.97 & 97.55 \\ 
\midrule
Proto-AWATT+LAS & 87.64$\pm${0.89} & 98.22$\pm${0.24} & 89.23$\pm${0.37} & 98.38$\pm$0.15 & 81.27$\pm${0.96} & 97.49$\pm$0.23 & 83.62$\pm${0.60} & 97.95$\pm$0.24 \\ 
Proto-AWATT+LCL & 87.44$\pm$0.88 & 98.09$\pm$0.25 & 89.08$\pm$0.73 & 98.31$\pm$0.11 & 81.11$\pm${0.95} & 97.49$\pm${0.13} & 83.29$\pm${0.94} & 97.93$\pm${0.22} \\
Proto-AWATT+SCL & 86.76$\pm${0.51} & 97.84$\pm${0.23} & 88.69$\pm${0.97} & 98.28$\pm${0.19} & 80.33$\pm${1.08} & 97.34$\pm${0.13} & 83.02$\pm${0.65} & 97.89$\pm$0.54 \\
\midrule
\multicolumn{9}{c}{\textit{FewAsp(multi)}}\\
\midrule
Proto-AWATT & 71.72 & 91.45 & 77.19 & 93.89 & 58.89 & 89.80 & 66.76 & 92.34 \\ 
\midrule
Proto-AWATT+LAS & 72.63$\pm${0.88} & 92.29$\pm${0.54} & 78.06$\pm${0.43} & 94.12$\pm${0.24} & 61.50$\pm${0.30} & 90.81$\pm${0.21} & 67.30$\pm${0.51} & 92.84$\pm${0.25} \\ 
Proto-AWATT+LCL & 72.61$\pm${0.82} & 92.08$\pm${0.36} & 77.78$\pm${0.92} & 93.92$\pm${0.53} & 60.42$\pm${0.47} & 89.93$\pm$0.37 & 67.12$\pm${0.78} & 92.52$\pm${0.67} \\ 
Proto-AWATT+SCL & 72.03$\pm$0.31 & 91.78$\pm$0.17 & 77.39$\pm${0.86} & 93.90$\pm${0.35} & 59.42$\pm${0.97} & 89.89$\pm$0.31 & 66.89$\pm${0.85} & 92.38$\pm$0.72 \\
\bottomrule
\end{tabular}}
\caption{Ablation study over two main components of LDF-AWATT. Besides, we also report the ablation result of Proto-AWATT+LCL. We report the average performance and standard deviation over 5 runs.}
 \label{ap:ablation_study_a}%
\end{table*}




\subsection{Error Analysis}\label{error}


To analyze the limitations of our framework, we randomly sample 100 error cases by \emph{LDF-AWATT} from the {\tt FewAsp} dataset, and roughly classify them into two categories. Table \ref{tab:error} shows the proportions and some representative examples for each category. The primary category is Complex, which includes examples that require deep comprehension to be understood. As shown in example (1), the word fragment ``{\it Chandler downtown Serrano}'' related to restaurant\_location appears no more than five times in the dataset, the low frequency of those expressions makes it hard for our model to capture their patterns, so it is really challenging to give a right prediction. The second category is no obvious clues, which includes examples with insufficient information. As shown in example (2), the sentence is very short and unable to provide abundant information to predict the true label.

Through the error analysis, we can conclude that although current models have achieved appealing progress, there are still some complicated sentences beyond their capabilities. There ought to be more advanced natural language processing techniques to further address them.


\begin{table*}[t]
\centering
\scalebox{0.80}{
\begin{tabular}{p{2.6cm}|l|p{6cm}|p{3.5cm}|l}
\toprule
{Category} & Proportion & Example & True Label & Predict Label\\
\midrule
\multirow{4}{*}{Complex} & \multirow{4}{*}{41\%} & (1) fast forward to december 2014, we have a company gathering in one of the many banquet rooms at the chandler downtown serrano. & \multirow{4}{*}{restaurant\_location}  & \multirow{4}{*}{room\_interior \xmark} \\
\midrule
\multirow{2}{*}{No obvious clues} & \multirow{2}{*}{22\%} &  (2) overall, this is a great salon, and I will be back !  & procedure\_beauty\_nails experience\_wait & \multirow{2}{*}{salon\_interior\_room \xmark} \\
\bottomrule
\end{tabular}}
\caption{Error analysis of LDF-AWATT.}
\label{tab:error}%
\end{table*}

\end{document}